\documentclass[lettersize,journal]{IEEEtran}
\usepackage{amsmath,amsfonts}
\usepackage{algorithmic}
\usepackage{array}
\usepackage{subfig}
\usepackage{textcomp}
\usepackage{stfloats}
\usepackage{url}
\usepackage{verbatim}
\usepackage{graphicx}
\def\BibTeX{{\rm B\kern-.05em{\sc i\kern-.025em b}\kern-.08em
    T\kern-.1667em\lower.7ex\hbox{E}\kern-.125emX}}
\usepackage{balance}

\usepackage{cite}
\usepackage{amsmath,nicefrac}
\usepackage{bm}
\usepackage{amsthm}
\usepackage{amssymb}
\usepackage{amsfonts}
\usepackage{mathtools}
\usepackage{graphicx}
\usepackage[colorlinks=true, allcolors=blue]{hyperref}
\usepackage{bbm}
\usepackage{url}
\usepackage{color}
\usepackage{graphicx}
\usepackage{caption}
\usepackage{algorithmic}
\usepackage[ruled,vlined]{algorithm2e}
\usepackage{verbatim}
\usepackage[pdftex,dvipsnames]{xcolor}  
\usepackage[colorinlistoftodos,prependcaption]{todonotes}
\usepackage{wrapfig} 
\usepackage[T1]{fontenc}
\newcommand{\rev}[1]{ \textcolor{black}{#1}}
\newcolumntype{P}[1]{>{\centering\arraybackslash}p{#1}}

\begin{document}
\title{A Cost-Sensitive Transformer Model for Prognostics Under Highly Imbalanced Industrial Data}

\author{Ali Beikmohammadi,
Mohammad Hosein Hamian, 
Neda Khoeyniha, 
Tony Lindgren,
Olof Steinert,
and Sindri Magn\'usson 
\thanks{A. Beikmohammadi, T. Lindgren, and S. Magn\'usson are with the Department of Computer and System Science, Stockholm University, 11419 Stockholm, Sweden (e-mail: beikmohammadi@dsv.su.se; tony@dsv.su.se; sindri.magnusson@dsv.su.se).}
\thanks{M. H. Hamian is with the Faculty of Computer Engineering, K. N. Toosi University of Technology, Tehran, Iran (e-mail: hamian@email.kntu.ac.ir).}
\thanks{N. Khoeyniha is with the Faculty of Management, University of Tehran, Tehran, Iran (e-mail: neda.khoeyniha@ut.ac.ir).}
\thanks{T. Lindgren and O. Steinert are with the Scania CV, Strategic Product Planning and Advanced Analytics, S\"odert\"alje, Sweden (e-mail: olof.steinert@scania.com).}
}


\maketitle

\begin{abstract}
\rev{
The rapid influx of data-driven models into the industrial sector has been facilitated by the proliferation of sensor technology, enabling the collection of vast quantities of data. However, leveraging these models for failure detection and prognosis poses significant challenges, including issues like missing values and class imbalances. Moreover, the cost sensitivity associated with industrial operations further complicates the application of conventional models in this context. 
This paper introduces a novel cost-sensitive transformer model developed as part of a systematic workflow, which also integrates a hybrid resampler and a regression-based imputer.
After subjecting our approach to rigorous testing using the APS failure dataset from Scania trucks and the SECOM dataset, we observed a substantial enhancement in performance compared to state-of-the-art methods. Moreover, we conduct an ablation study to analyze the contributions of different components in our proposed method.
Our findings highlight the potential of our method in addressing the unique challenges of failure prediction in industrial settings, thereby contributing to enhanced reliability and efficiency in industrial operations.
}
\end{abstract}

\section{Introduction}
\IEEEPARstart{P}{rognostics} pertains to predicting the future state of a given component or system, taking into account its present and past conditions. 
This capability enables proactive maintenance and planning, mitigating the risk of critical failures and subsequently reducing downtime and associated costs \cite{22heng2009rotating}.
This concept has witnessed rapid growth in recent years, as it plays a vital role in assessing and forecasting the behavior and reliability of physical systems. \rev{Prognostics is applied in many domains such as aerospace engineering \cite{1galanopoulos2021health}, renewable energy \cite{4aizpurua2022hybrid}, autonomous driving \cite{5raouf2022sensor}, nuclear power and chemical plants \cite{7hu2012ensemble}, robotics \cite{ 9qiao2018monitoring}, railway systems \cite{wang2016bilevel}, self-organizing network \cite{han2022evolutionary}, fraud detection \cite{zhu2020optimizing}, and civil infrastructure \cite{10vega2022diagnosis}. }


The approaches designed for prognostics fall into three categories based on their underlying principles: data-driven, model-based, or hybrid approaches \cite{11guo2019review}. Among these categories, the data-driven approach, which relies on the availability and completeness of data, has currently garnered significant interest due to the rapid growth of sensor technology, network capacity, and storage capacity, which have facilitated its application. 
Additionally, the rise of machine learning (ML) techniques, particularly deep learning (DL), has further empowered data-driven models by harnessing extensive datasets to discern the relationships between system inputs and outputs based on historical data.
These DL techniques have the potential to enhance the precision, resilience, and efficiency of data-driven approaches while also offering deeper insights into the root causes and mechanisms behind degradation and failure.
However, there exist certain challenges associated with working with data collected for prognostic purposes that can impede the achievement of satisfactory results when employing DL models.


The first critical challenge encountered when collecting data through sensors and communication is the presence of missing values, which poses a significant challenge in the development of data-driven prognostic models. These missing values can substantially impact the accuracy and reliability of predictive outcomes. As a result, it becomes imperative to address missing values appropriately before employing any ML or data mining techniques. Various methods are available for handling missing data, including deletion, imputation, or incorporating them as a feature \cite{13pratama2016review}. Each of these methods carries its own set of advantages and disadvantages. The selection of an appropriate method hinges on factors such as the research objectives, data characteristics, and analytical assumptions.

The second significant challenge faced by data-driven prognostic models is class imbalance, which arises due to the fact that the occurrence of failure events is much rarer compared to other events within the dataset. This imbalance can pose difficulties for ML models since they often exhibit a bias towards the majority class, leading to suboptimal predictions and biased results. To address this issue, imbalanced data processing techniques have been developed to enhance the performance and accuracy of ML models when dealing with datasets exhibiting skewed class distributions in the target variable, taking into account the extent of imbalance, the model type, and the specific problem domain \cite{14haixiang2017learning}. 
Common techniques encompass resampling methods \cite{15pereira2021toward}, class weighting approaches \cite{17liu2011class}, and ensemble learning techniques \cite{18malhotra2020handling}. However, it is important to note that these methods may have limitations and challenges, including potential information loss, increased computational complexity, or the need for parameter tuning.

The third influential challenge encountered in the development and application of data-driven prognostic models relates to the presence of cost sensitivity within the problem domain. Cost sensitivity is a crucial factor in numerous real-world applications. In these scenarios, the costs and consequences associated with different types of errors are not uniform, and as a result, evaluating classifier performance solely based on accuracy is inadequate. For instance, in fields such as medical diagnosis \cite{19gan2020integrating}, fraud detection \cite{20nami2018cost}, spam filtering \cite{zhou2014cost}, and APS failure detection \cite{36taghandiki2023minimizing}, the cost associated with a false negative (missing a positive case) is often considerably higher than that of a false positive (misclassifying a negative case). Consequently, it is imperative to develop ML algorithms capable of taking into account misclassification costs and minimizing the overall cost rather than prioritizing accuracy alone.


\rev{
\textbf{Contribution:} In this paper,  we embark on addressing challenges such as imbalanced data, missing values, and cost sensitivity within prognostic tasks, introducing a novel workflow tailored to these specific hurdles. 
To provide further clarity, we very firstly develop a Transformer model from the ground up, equipping it with the capability to effectively learn from intricate datasets. Additionally, we employ a focal loss, inspired by techniques from computer vision (CV), to formulate a loss function tailored for managing cost-sensitive problems. 
Moreover, within our methodology, we pioneer the use of a hybrid resampling technique in the prognostic domain. This innovative approach combines a modified version of the SMOTE algorithm, employing an SVM algorithm to pinpoint samples for generating new synthetic data points, and the Repeated Edited Nearest Neighbor (Repeated ENN) method to address imbalanced data concerns.
Through this novel approach, we seek to pave the way for more effective and data-driven approaches to predictive maintenance in various industrial sectors.
Finally, we evaluate the effectiveness of our approach in terms of different metrics on the highly imbalanced, cost-sensitive APS failure dataset derived from Scania trucks and the SECOM dataset, demonstrating that our results surpass those of all existing state-of-the-art methods. 
We conduct an ablation study to dissect and showcase the individual contributions of each component of our proposed method, providing further insights into its overall efficacy.
}

The remainder of this paper is organized as follows. In Section \ref{lit}, we review the existing related studies. In Section \ref{met}, we present our proposed method in detail. In Section \ref{exp}, we report the results and experiments of our proposed method on the APS dataset. 
\rev{Additionally, we include results on the SECOM dataset in Appendix.} 
In Section \ref{conc}, we conclude the paper and discuss the future directions.

\section{Literature review} \label{lit} 
\noindent While our study introduces pioneering elements such as the utilization of the Transformer model, the incorporation of cost sensitivity through Focal Loss, and the application of innovative undersampling and oversampling strategies, we broaden the scope of this section to encompass both advancements within the prognostic domain and techniques developed to address the aforementioned challenges across various contexts beyond prognostics. 

\subsection{\rev{Machine Learning in Prognostic Prediction}}

\noindent Prognostic is one of the domains where ML and DL techniques find their application, learn from large amounts of data, and make predictions based on complex patterns. In the realm of traditional ML algorithms, several popular models have been extensively proposed and compared in terms of their performance across various metrics. Notably, Naive Bayes, K-Nearest Neighbors (KNN), Support Vector Machines (SVM), Random Forest, and Gradient Boosted Trees have garnered significant attention \cite{30rafsunjani2019empirical, 012chen2020direct, 38ranasinghe2019methodology, 40akarte2018predictive, 015yang2020lifespan}.
In this context, it is worth noting that both gradient boosting and random forest algorithms have demonstrated better performance compared to their counterparts. 
In contrast, DL techniques have been suggested for the purpose of prognosis prediction and have exhibited superior performance compared to classical ML methods. Among the DL methods, long short-term memory (LSTM) networks \cite{31oh2023quantum, 27ke2022deep, 018liu2019deep}, along with Bayesian deep learning (BDL)\cite{019peng2019bayesian} and other DL networks specifically designed for this purpose \cite{36taghandiki2023minimizing, 37oh2020imbalanced, 41sun2023robust}, have emerged as particularly suitable for prognosis prediction due to their capacity to effectively handle uncertainty and dependencies inherent in the data. 

However, there is a recognized requirement for a more sophisticated network that can efficiently attend to the inter-relationships among input features while also detecting their temporal dependencies. The transformer model \cite{vaswani2017attention} has emerged as the best solution for this type of data. One of the key advantages of transformers is their ability to handle long-range dependencies in sequential data. This is made possible by the self-attention mechanism, which allows transformers to compute similarity scores between all pairs of features in the input sequence, regardless of their distance. Consequently, transformers are able to weigh the importance of each feature based on these similarity scores, enabling them to capture semantic and syntactic relationships between features.
\rev{Building upon these advancements, we, in this paper, break new ground by introducing the Transformer model for the first time in the domain of predictive maintenance.}

\subsection{Dealing with Missing Values Challenge}
 \noindent Prognostic problems often face the challenge of missing values in the data, a common problem in data analysis that can affect the validity and reliability of the obtained outcomes.  Numerous approaches have been suggested to address the presence of missing values, encompassing imputation techniques that depend on probabilistic models or ML models to complete the gaps 
 \cite{025chiu2022missing}. 
 Some of the widely used and effective statistical imputation methods are expectation maximization \cite{027zhang2014expectation}, Gaussian mixture model \cite{030przewikezlikowski2020estimating}, multiple imputations by chained equations (MICE) \cite{031pauzi2021comparison}, Bayesian imputation \cite{032zhang2015using}. 
 On the other hand, ML-based imputation methods include random forest \cite{033tang2017random}, CKNNI \cite{034jiang2015cknni}, SVM \cite{035mallinson2003imputation}, and various types of clustering \cite{036khan2021missing, 037zhang2008missing}, all of which have demonstrated strong performance. 
 
 Considering the various aspects of statistical and ML-based imputation methods, Bayesian Ridge Regression has demonstrated superior performance in terms of accuracy and computational efficiency compared to the other approaches, particularly when dealing with datasets characterized by high dimensionality or intricate patterns, as evidenced by multiple studies \cite{039pereira2020vae, 040m2020cbrl}. To deal with missing values dealing with prognostic datasets, we aim to apply this powerful approach.
\rev{In the pursuit of adeptly managing missing values within prognostic datasets, we embark on an endeavor to explore the efficacy of this formidable technique within this domain, examining its suitability and performance.}

\subsection{Dealing with Imbalanced Data Challenge}
\noindent Dealing with imbalanced data is a frequent and difficult problem in data analysis and ML, which can impact the effectiveness and precision of the models. 
One of the most effective ways to address the problem of imbalanced data handling is resampling, which involves changing the distribution of dataset \cite{14haixiang2017learning}.
Resampling methods can be categorized into three primary approaches: undersampling, aimed at reducing the volume of the majority class; oversampling, intended to augment the size of the minority class; and hybrid methods, which encompass a combination of both undersampling and oversampling techniques. 
 
 Several undersampling methods have been proposed, such as parallel selective sampling \cite{044d2015parallel}, near miss \cite{045tanimoto2022improving}, clustering-based undersampling 
 \cite{048onan2019consensus}, ENN \cite{wilson1972asymptotic} 
 , weighted undersampled SVM \cite{kang2017distance}, 
 and Repeated ENN \cite{tomek1976experiment} to overcome imbalance data issue. 
Among these undersampling methods, Repeated ENN has been known as a powerful technique to improve the performance of imbalanced classification problems in various domains \cite{tomek1976experiment}. 
 On the other hand, oversampling methods have been proposed to address the imbalanced data problem by generating synthetic samples for the minority class, such as SWIM \cite{053bellinger2020framework}, SMOTE \cite{chawla2002smote}, Reverse-SMOTE \cite{051das2020oversampling}, Borderline-SMOTE \cite{052han2005borderline}, and SVM-SMOTE \cite{nguyen2011borderline}. 
 SVM-SMOTE can effectively handle imbalanced data with high dimensionality, complex patterns, and overlapping classes.

\rev{In line with prior research findings, as indicated by numerous studies \cite{054choirunnisa2018hybrid, 055shamsudin2020combining}, which suggest that the concurrent application of undersampling and oversampling techniques can yield a more informative dataset and enhance classification performance, our paper endeavors to pioneer a hybrid approach that combines SVM-SMOTE and Repeated ENN. Notably, this combination represents a novel contribution, as no prior study has explored this particular hybridization.}

 \subsection{Dealing with Cost Sensitivity Challenge}
\noindent In ML classification, cost sensitivity is an essential factor that influences the performance of the model. By minimizing the expected cost rather than maximizing the accuracy, the model can better handle the diverse outcomes. A prevalent method for integrating cost into the learning process involves the adaptation of the training algorithm of the model, commonly referred to as algorithm modification. This can be accomplished by implementing a loss function that is weighted or by utilizing a cost matrix that assigns greater penalties to errors that are more expensive \cite{056sun2007cost}. Several methods have been proposed to develop cost-aware ML algorithms, such as near-bayesian SVM \cite{057datta2015near}, cost-sensitive KNN \cite{058qin2013cost}, and cost-sensitive neural network \cite{059zhou2005training}. 

One of the possible modifications of DL-based methods to account for cost sensitivity is to use Focal Loss \cite{lin2017focal}, primarily designed for CV applications, as the loss function to update the model parameters. 
\rev{To the best of our knowledge, no study has applied Focal Loss to address cost sensitivity in DL models for prognostic tasks.}

\section{\rev{Proposed Method}} \label{met}
\noindent In this section, we present the methodology underpinning our work, which consists of four fundamental components: handling missing values, mitigating the effects of imbalanced data, the Transformer model, and the creation of a cost-sensitive loss function.
The first two components pertain to the preprocessing steps that involve converting raw data into a format suitable for our model's input. 
The third component is our model's architecture, which is built upon a deep Transformer neural network featuring multiple layers and activation functions. 
Lastly, the fourth component deals with the criteria used to assess the disparity between our model's predictions and the actual ground truth labels. We aim to couple this assessment with cost sensitivity. 
Detailed descriptions of each component are provided in the subsequent sections.

\subsection{Handling Missing Values} \label{metMis}

\noindent We have devised a two-stage approach to enhance the efficiency and quality in dealing with missing data. 

\subsubsection{Eliminating Features with a Large Number of Missing Values} \label{metMisEl}
\noindent To address the issue of missing values, our initial strategy involves the removal of attributes exhibiting a significant number of missing entries. 
Attributes with a high volume of missing values often lack enough information for precise imputation, resulting in potentially unreliable imputed values.
Through the elimination of these particular features, we reduce the burden of imputation and concentrate on the remaining data, which offers more comprehensive information. This, in turn, enhances the overall quality of the imputed dataset.
\rev{Note that the impact of removing features with missing entries varies based on the dataset, and the decision to eliminate features depends on the specific characteristics and problem requirements.}

\subsubsection{\rev{Bayesian Ridge Regression-based Imputation for the Remaining Missing Values}}\label{metMisBay}
\noindent 
Bayesian Ridge Regression offers a probabilistic framework for imputing values, proving particularly advantageous when working with incomplete datasets \cite{39tipping2001sparse}. This is of paramount importance as it permits the quantification of our confidence levels in the imputed values, thus supplying valuable information for subsequent analyses.

In Bayesian Ridge Regression, the model assumes a prior distribution over the model parameters and updates this prior based on the observed data to obtain a posterior distribution over the parameters. The prior distribution for the parameters is typically chosen to be a Gaussian distribution with mean 0 and a precision (inverse variance) hyperparameter, denoted by $\alpha$. The likelihood of the data is assumed to be normally distributed with a mean given by the linear regression model and a precision hyperparameter, denoted by $\lambda$.

The linear regression model can be represented as:
\begin{equation}
y = Xw + \epsilon.
\end{equation}
In this equation, $y$ denotes the target variable, $X$ stands for the design matrix of features, $w$ represents the vector of regression coefficients (parameters) to be estimated, and $\epsilon$ signifies the error term. It is worth noting that the error term is assumed to follow a normal distribution with a mean of 0 and precision equal to $\lambda^{-1}$.

The Bayesian Ridge Regression model places a Gaussian prior on the parameters $w$:
\begin{equation}
P(w) = \mathcal{N}(0, \alpha^{-1}I),
\end{equation}
where $P(w)$ is the prior distribution over $w$. $\alpha$ is the precision (inverse variance) of the prior; $I$ is the identity matrix.

The likelihood of the data, given the model and the parameters, is assumed to be Gaussian:
\begin{equation}
P(y | X, w, \lambda) = \mathcal{N}(Xw, \lambda^{-1}I).
\end{equation}

The goal is to compute the posterior distribution of the parameters $w$ given the data and the prior:
\begin{equation}
P(w | X, y, \alpha, \lambda) \propto P(y | X, w, \lambda) \cdot P(w).
\end{equation}

Utilizing Bayesian Ridge Regression, we adopt an iterative approach to address missing values, treating them as missing at random. This process commences with the initial imputation of missing values, employing simple mean estimates. Subsequently, we iteratively enhance these estimates by incorporating our regression model, which relies on the observed values of other features.


\subsection{Addressing Imbalanced Data} \label{metIm}
\noindent To address imbalanced data, we employ a combination of oversampling and undersampling techniques to rebalance the class distribution. In our proposed two-step action, we suggest using the SVM-SMOTE method \cite{nguyen2011borderline} and the Repeated ENN method \cite{tomek1976experiment}. 

\subsubsection{\rev{Oversampling by SVM-SMOTE}} \label{metImSVM}
\noindent SVM-SMOTE is an effective oversampling technique that combines the Synthetic Minority Over-sampling Technique (SMOTE) \cite{chawla2002smote} with SVMs. This approach aims to generate synthetic samples for the minority class by interpolating between existing minority class samples and their nearest neighbors. The primary goal is to ensure that these synthetic samples are situated in regions of the feature space most likely to belong to the minority class. This strategy takes into account the local distribution of the minority class, enhancing its ability to capture the underlying data structure.

Utilizing SVMs, which excel at defining decision boundaries in high-dimensional spaces, SVM-SMOTE has the potential to produce high-quality synthetic samples. The SVM-SMOTE technique leverages an SVM algorithm to identify minority class samples that are either prone to misclassification or are situated close to the decision boundary. Subsequently, like SMOTE, it generates synthetic samples by linearly interpolating between these selected samples and their nearest neighbors within the minority class. The mathematical formula used for generating a synthetic sample is expressed as:
\begin{equation}
 s = x + \beta (x' - x)
\end{equation}
Here, $s$  represents the synthetic sample, $x$ is the chosen minority class sample, $x'$  is its nearest neighbor within the minority class, and $\beta$ is a randomly generated number ranging from 0 to 1. The pseudocode of this algorithm is expressed in Algorithm \ref{SVMSMOTE}.

\begin{algorithm}[h]
\SetAlgoNlRelativeSize{0}
\SetKwInput{KwData}{Input}
\SetKwInput{KwResult}{Output}
\KwData{Training dataset}
\KwResult{Oversampled dataset}
Initialize empty oversampled dataset\;
Fit an SVM classifier on the training dataset\;
Identify support vectors from the minority class\;
\ForEach{minority sample $x$ in the training dataset}{
    Find $k$-nearest neighbors among the minority class\;
    Randomly select a neighbor $x'$\;
    Generate a synthetic sample $s$ by calculating $s=x+\beta(x'-x)$\;
    Add the synthetic sample $s$ to the oversampled dataset\;
}
Combine the original training dataset with the oversampled dataset\;
\caption{SVM-SMOTE}
\label{SVMSMOTE}
\end{algorithm}

\subsubsection{\rev{Undersampling by Repeated ENN}} \label{metImENN}
\noindent Edited Nearest Neighbors (ENN) is an undersampling technique that identifies and eliminates potentially noisy samples by iteratively examining the nearest neighbors of each sample \cite{wilson1972asymptotic}. These samples are removed if the majority of their nearest neighbors belong to a different class, known as noisy samples. ENN is particularly useful when dealing with datasets that exhibit overlapping regions between classes.
To enhance the noise removal process, there's a variant called Repeated ENN \cite{tomek1976experiment}. 

Similar to ENN, Repeated ENN's primary objective is the removal of noisy samples, primarily from the majority class, all while ensuring the preservation of the minority class's decision boundary. However, it distinguishes itself by accomplishing this goal through the repetitive application of the ENN method.
The process continues until one of the following conditions is met: no more samples can be removed, a maximum number of iterations is reached, or one of the majority classes either becomes a minority class or disappears due to undersampling. The pseudo-code for the Repeated ENN method is provided in Algorithm \ref{RENN}:

\begin{algorithm}[h]
\SetAlgoNlRelativeSize{0}
\SetKwInput{KwData}{Input}
\SetKwInput{KwResult}{Output}
\KwData{Training dataset, Maximum number of iterations}
\KwResult{Undersampled dataset}
Initialize undersampled dataset by copying training dataset\;
\Repeat{The maximum number of iterations is reached \algorithmicor ~No more samples can be removed from the undersampled dataset \algorithmicor ~One of the majority classes becomes a minority class}{
\ForEach{majority sample $x$ in the undersampled dataset}{
    Find $k$-nearest neighbors among the undersampled dataset\;
    \If{the majority of neighbors' classes is different from $x$ class}{
        Remove the sample $x$ from the undersampled dataset\;
    } } }
\caption{Repeated ENN}
\label{RENN}
\end{algorithm}

\subsection{Model Architecture: The Transformer Neural Network} \label{metModel}

\begin{figure*}
    \centering
    \includegraphics[width=\textwidth]{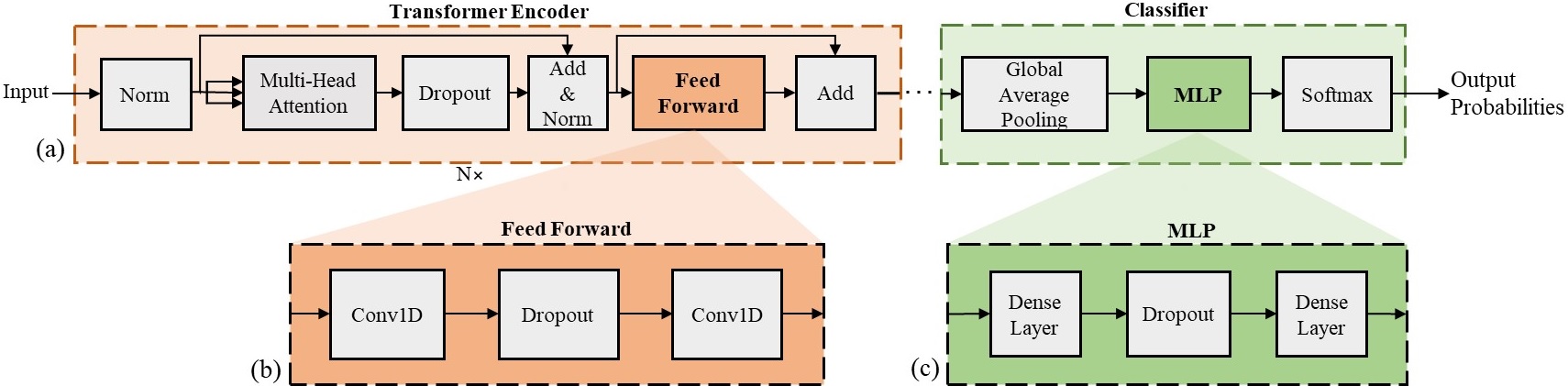}
    \caption{ The visual representation of our proposed model's design (a) Overall architecture of our model, (b) Details of the feed forward network, (c)  A detailed insight into the proposed MLP module.}
    \label{ourmodel}
\end{figure*}

\noindent The Transformer, a groundbreaking DL architecture introduced by Vaswani et al. in \cite{vaswani2017attention}, has revolutionized natural language processing (NLP) and ML by utilizing a self-attention mechanism, allowing it to capture contextual relationships in data efficiently. 
Transformers have found widespread applications in various domains, including NLP tasks like machine translation, text summarization, and sentiment analysis \cite{naseem2020transformer}. They have also been adapted for CV tasks like object detection and image captioning \cite{cornia2020meshed}. 
The key reasons for their success are their ability to handle long-range dependencies in data, scalability to large datasets, and parallelization, making them a vital tool in the advancement of artificial intelligence research.

In line with the transformative impact of the Transformer architecture, we pioneer its application in the realm of predictive maintenance, specifically focusing on prognostics and failure detection. 
Our work represents a significant departure from traditional ML and DL methods in this domain, leveraging the self-attention mechanism and contextual learning capabilities of Transformers to enhance predictive accuracy, enable earlier detection of potential failures, and ultimately optimize equipment reliability and operational efficiency. 

\rev{To provide a detailed description, Fig. \ref{ourmodel} visualizes the architecture of our from scratch-designed model. 
Please refer to Section \ref{ModelParamSec} and Appendix to see the hyperparameters that have been used in this model.
Our model comprises two primary components: a Transformer encoder utilized for feature extraction and a classifier responsible for making predictions. We stack $N$ encoder block to define a Transformer encoder in the model.
Each encoder block, in turn, can be divided into two distinct subcomponents: one focused on normalization and attention and another dedicated to feed forward operations.}

In detail, the first subcomponent consists of:
\begin{itemize}
    \item \textit{Normalization layer}: This layer normalizes both the input and the sum of the input and the output of the attention mechanism, helping stabilize training. This technique reduces the variance of the layer outputs and makes the network less sensitive to the scale of the parameters. Formally, this layer is defined as:
    \begin{equation}
        \text{LayerNorm}(x+ \text{Sublayer}(x)),
    \end{equation}
    where $x$ is the input vector, $\text{Sublayer}(x)$ is the output of a sub-layer (such as multi-head attention or feed forward network), and \text{LayerNorm} is the layer normalization function, which is defined as follows:
    \begin{equation}
        \text{LayerNorm}(x) = \zeta \frac{x-\mu}{\sigma}+\beta,
    \end{equation}
    where $\mu$ and $\sigma$ are the mean and standard deviation of $x$, respectively, and $\zeta$ and $\beta$ are learnable parameters that scale and shift the normalized vector.
    
    \item \textit{Multi-Head Attention}: This layer performs multi-head self-attention on the input sequence, allowing the model to compute a weighted sum of a set of vectors, where the weights are determined by the similarity between each pair of vectors. The attention layer can be used to capture the dependencies between the input and output tokens, regardless of their distance in the sequence. This layer operates based on the following equation:
    \begin{equation}
        \text{Attention}(Q,K,V) = \text{softmax}(\frac{QK^T}{\sqrt{d_k}})V,
    \end{equation}
    where $Q$, $K$, and $V$ are the query, key, and value matrices, respectively, and $d_k$ is the dimensionality of the keys. \text{softmax(.)} is the softmax function, which is used to convert a vector of real numbers into a probability distribution. It is defined as follows:
    \begin{equation}
       \text{softmax}(x)_i = \frac{e^{x_i}}{\sum_{j=1}^{n} e^{x_j}}
    \end{equation}
    Here, $n$ is the number of elements in the input vector.
    
    The intuition behind the attention layer is that each query vector can be seen as a question, and each key vector can be seen as a possible answer. The dot product between a query and a key measures how well they match, and the softmax function normalizes the scores to obtain a probability distribution. The value matrix contains the information that the model wants to retrieve, and the weighted sum of the values gives the output vector.

    \item \textit{Dropout}: Dropout is applied to the output of the attention mechanism for regularization and to prevent overfitting. It randomly sets a fraction of input units to zero during each forward pass. The mathematical representation of dropout is as follows:
    \begin{equation}
        \text{Dropout}(x, p) = \begin{cases}
           0 & \text{with probability } p \\
           \frac{x}{1 - p} & \text{otherwise}
        \end{cases}
    \end{equation}
    Here, $x$ represents the input value, and $p$ is the dropout probability.
    
    \item \textit{Residual Connection}: The output of the dropout layer is added to the original input, creating a residual connection.  Another residual connection applies to the output of the feed forward network to the result of the previous normalization layer.
    
\end{itemize}

The resulting tensor of the first subcomponent is then sent to the feed forward subpart, which itself, as shown in Fig. \ref{ourmodel}(b), consists of two 1D convolutional (Conv1D) layers with ReLU activation functions and dropout.
Conv1D uses a sliding window to combine multiple input values to produce an output value, which can be used to extract local features from the input sequence and to reduce the dimensionality of the vectors.
ReLU is the rectified linear unit activation function, which is defined as follows:
\begin{equation}
   \text{ReLU}(x) = \begin{cases} 
      0 & \text{if } x < 0 \\
      x & \text{if } x \geq 0 
   \end{cases}
\end{equation}

After the Transformer encoder, in the classifier component, we first apply a global average pooling layer to reduce the spatial dimension.
This layer computes the average value of each vector along a specified axis, which can be used to obtain a fixed-length representation of the input sequence and to reduce the number of parameters, overfitting, and computational cost of the model.
The intuition behind this layer is that it can capture the global information of the input sequence and ignore the local variations and noise in the data, resulting in a compact and meaningful representation of the data.

The output of the global average pooling layer is passed through a multi-layer perceptron (MLP), which consists of two dense (fully connected) layers with ReLU activation functions, followed by dropout layers for regularization (shown in Fig. \ref{ourmodel}(c)), and an output layer with a single neuron with a sigmoid activation function to squash real-valued number into the range [0, 1] (i.e., $\text{sigmoid}(x) = \frac{1}{1 + e^{-x}}$). 
The dense layers increase the dimensionality of the vectors and introduce more non-linearity into the model, formulated as:
\begin{equation}
    \text{Dense}(x) =  \text{ReLU}(xW+b),
\end{equation}
where $x$ is the input matrix, $W$ is the weight matrix, and $b$ is the bias vector.

\subsection{Cost-Sensitive Loss Function Design} \label{metCost}
\noindent In order to train our model effectively, it is essential to establish an appropriate loss function. In the context of predictive maintenance, it's common for the cost associated with failing to detect a positive case to be significantly greater than that of incorrectly classifying a negative case (i.e., $C_{FN} >> C_{FP}$). Therefore, we contend that utilizing a symmetric loss function such as cross entropy may not effectively account for the costs of misclassification. Instead, our objective is to minimize the overall cost as a priority rather than prioritizing accuracy alone.

To mitigate the cost sensitivity inherent in our problem, we have adopted the Focal Loss as our chosen loss function. Focal Loss is a specialized loss function primarily crafted to tackle the issue of class imbalance in DL models, with a specific emphasis on applications like object detection tasks in CV \cite{lin2017focal}.
This loss function builds upon the foundation of cross entropy loss but introduces a novel concept: it assigns varying degrees of importance to different samples during the training process. It accomplishes this by assigning a higher weight to what are termed "hard" examples while correspondingly reducing the weight for easily classified examples. Essentially, Focal Loss reduces the loss for well-classified instances, enabling the model to concentrate more on enhancing its performance with challenging samples.

Mathematically, we begin with the cross entropy loss, as the Focal Loss is built upon it. The mathematical representation of cross entropy loss is as follows:
\begin{align}
    &\text{CrossEntropyLoss}(p_t) =  -\log(p_t), \nonumber \\
    &p_t = \begin{cases} 
      p & \text{if } y = 1 \\
      1-p & \text{otherwise}  
   \end{cases}
\end{align}
where $p$ is the predicted probability vector for the positive class, and $y$ is the ground-truth class vector (usually a one-hot vector). For binary classification, the cross entropy loss can be simplified as follows:
\begin{align}
    \text{BinaryCrossEntropyLoss}(p, y) = & -y\log(p) \nonumber \\
   & -(1-y)\log(1-p).
\end{align}

The Focal Loss modifies the cross entropy loss by adding a modulating factor $(1 - p_t)^\gamma$, resulting in:
\begin{equation}
    \text{FocalLoss}(p_t) =  -\alpha_t(1-p_t)^\gamma\log(p_t).
\end{equation}
This modulating factor reduces the loss value for well-classified examples (when $p_t$ is close to 1) and increases the loss value for misclassified examples (when $p_t$ is close to 0). The focusing parameter $\gamma$ controls the emphasis on the hard examples, with larger values giving more focus to challenging instances. $\alpha_t$ serves as a balancing factor, often set as a constant (e.g., inversely proportional to class frequency).

The Focal Loss for binary classification can be expressed as:
\begin{align}
\text{BinaryFocalLoss}(p, y) =& -y\alpha(1-p)^\gamma\log(p) \nonumber \\ 
&-(1-y)(1-\alpha)p^\gamma\log(1-p)   
\end{align}

In our particular context, characterized by the presence of cost-sensitive considerations, we deliberately assign a higher degree of importance to positive class samples, deeming them as representative of "hard" examples. This strategic maneuver significantly amplifies their influence during the learning process, effectively aligning our model's attention and optimization efforts with the specific requirements inherent in our cost-sensitive task.

To illustrate this strategic adjustment, consider the following example: When we set $\alpha$ to 0.95 and $\gamma$ to 1.5, we observe notable differences in the loss values. In a scenario where $y=1$ and $p=0.1$ (i.e., a false negative case), the Focal Loss yields an approximate value of 1.868. Conversely, in a situation where $y=0$ and $p=0.9$ (i.e., a false positive case), the Focal Loss returns an approximate value of 0.098. Remarkably, under the conventional cross entropy loss, both scenarios would yield identical loss values of 2.072. This example vividly underscores how our adoption of the Focal Loss function allows us to distinctly prioritize and address the nuanced challenges posed by false negatives and false positives within our cost-sensitive context.

\section{\rev{Case Study \MakeUppercase{\romannumeral 1}: Detecting APS failure}} \label{exp}
\noindent \rev{In this section, we begin by providing an overview of the problem's significance and its real-world relevance. Following this, we outline the dataset's source. We then define the performance metrics used to gauge our model's effectiveness. Moving on, we detail the preprocessing and cleansing steps applied to prepare the dataset for training. Subsequently, we cover essential aspects such as model architecture and training parameters. Ultimately, we showcase our experimental results, evaluate the model's performance based on predefined metrics, conduct a thorough comparative analysis with existing studies, and present the findings from an ablation study, dissecting and highlighting the individual contributions of each component in our proposed methodology.}
All source codes required for conducting and analyzing the experiments will be made available online\footnote{We share all source codes as supplementary materials.} upon acceptance of the paper.

\subsection{Problem Description}
\noindent Numerous ongoing research endeavors in prognostics are actively exploring the realms of failure detection, employing ML methodologies. 
As a case study, consider the potential ramifications of brake failure in a vehicle, which could precipitate severe accidents. Thus, it becomes imperative to detect and prevent such failures. 
In the context of heavy-duty vehicles like trucks and excavators, the efficiency of their braking systems assumes critical importance due to the substantial loads they need to bring to a halt. 
These vehicles rely on an air pressure system (APS) to power their brakes, necessitating a consistent supply of pressurized air for optimal functionality. 

\begin{figure}
    \centering
    \includegraphics[width=\linewidth]{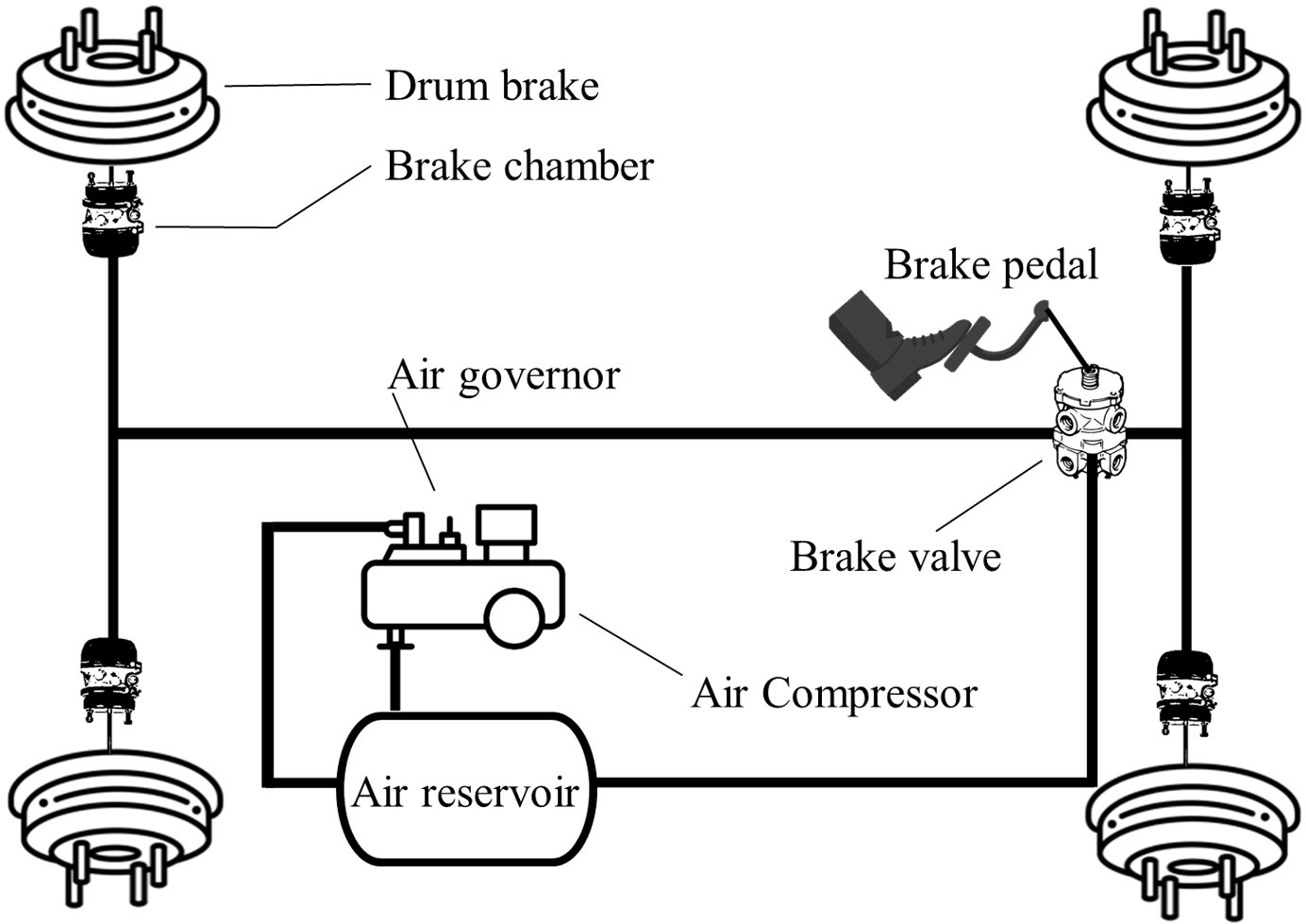}
    \caption{Air pressure system structure.}
    \label{aps}
\end{figure}

Fig. \ref{aps} provides an illustrative representation of the APS structure. As delineated in the figure, the air compressor initially compresses atmospheric air and regulates it to maintain an optimal pressure level via the air compressor governor. Subsequently, the compressed air is directed to the air reservoir. When a driver engages the brake pedal, the brake valve closes, and the compressed air stored in the air reservoir is deployed to apply pressure to the brake chamber, facilitating the braking process.
Any deviation in this system can compromise brake precision, potentially resulting in catastrophic accidents. 

\subsection{Dataset}
\noindent \rev{The dataset under examination, sourced from Scania Trucks and accessible via the UCI Machine Learning Repository \cite{dataset}, pertains to APS failures within heavy Scania trucks. This binary classification dataset encompasses two distinct categories: "positive," denoting trucks experiencing APS failures, and "negative," representing those without such issues.} 
A total of 76,000 instances and 170 attributes are included, featuring a cost metric associated with misclassifications. This dataset is well-suited for classification tasks, where the aim is to forecast and mitigate the overall cost of failures, relying on sensor-derived data. To protect proprietary information, attribute names have been anonymized. Moreover, the dataset has been partitioned into two subsets, with 60,000 records allocated for training and 16,000 designated for testing.

\rev{The dataset provides a cost metric for misclassification, notably discussed in \cite{dataset}, described by:}
\begin{equation} \label{cost}
    \text{Total Cost} = (10 \times FP) + (500 \times FN).
\end{equation}
In this metric, a false positive prediction (indicating a truck without APS failure is mistakenly predicted to have one) incurs a cost of \$10, while a false negative prediction (indicating a truck with APS failure is inaccurately predicted to be without it) carries a cost of \$500. The cumulative cost of a prediction model is calculated by summing the costs associated with individual instances.

\begin{figure}
    \centering
    \includegraphics[width=\linewidth]{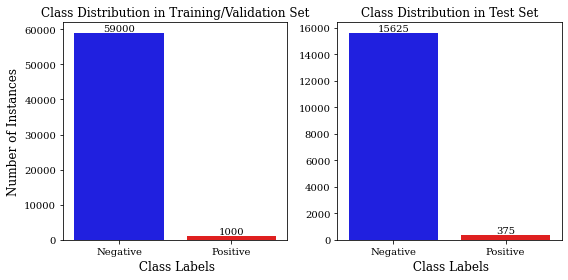}
    \caption{\rev{APS Failure at Scania Trucks dataset distribution.}}
    \label{datadistribution}
\end{figure}

Furthermore, as shown in Fig. \ref{datadistribution}, the dataset exhibits an imbalance issue, with a substantial overrepresentation of the negative class in comparison to the positive class. Specifically, the training set encompasses 59,000 records for the negative class and only 1,000 records for the positive class, yielding a class imbalance ratio of 59:1. Similarly, the test set comprises 15,625 records for the negative class and a mere 375 records for the positive class, resulting in a class imbalance ratio of 41:1. This skewed distribution poses a notable challenge for classification models.

\subsection{Performance Metrics} \label{PerformanceMetricsSec}
\noindent In this research, the primary focus is on the pivotal metric, the total cost, and the proportions of its constituent elements as defined in Equation \eqref{cost}, with our objective being its minimization. In addition to these critical metrics, we utilize the following evaluation criteria to appraise the performance of the proposed method and measure its effectiveness, concurrently conducting a comparative analysis with other research endeavors:
\begin{itemize}
\item True Positive ($TP$): corresponds to the complete count of correctly identified positive samples by the classifier, which indeed belongs to the positive class. 
\item False Positive ($FP$): signifies the total count of samples incorrectly classified as positive by the classifier when, in reality, their actual class label is negative. 
\item True Negative ($TN$): accounts for the comprehensive tally of samples correctly classified as negative by the classifier, and these samples genuinely belong to the negative class.
\item False Negative ($FN$): denotes the total count of samples inaccurately classified as negative by the classifier, while the truth is that their actual label is positive.
\item Accuracy $=\frac{TP + TN}{TP + TN + FP + FN}$
\item Precision $ = \frac{TP}{TP + FP}$
\item Sensitivity $ = \frac{TP}{TP + FN}$
\item Specificity $ = \frac{TN}{FP + TN}$
\item F1 Score $ = \frac{2TP}{2TP + FP+ FN}$
\item Negative Predictive Value: $NPV = \frac{TN}{TN + FN}$
\item False Discovery Rate: $FDR = \frac{FP}{FP + TP}$
\item False Positive Rate: $FPR = \frac{FP}{FP + TN}$
\item False Negative Rate: $FNR = \frac{FN}{FN + TP}$
\item False Omission Rate: $FOR = \frac{FN}{FN + TN}$
\end{itemize}
These metrics can be categorized into two distinct subsets based on their optimization goals: the first group comprises metrics where higher values are preferable, including TP, TN, Accuracy, Precision, Sensitivity, Specificity, F1 Score, and NPV. Conversely, the second group consists of metrics where lower values are desirable, which encompasses FDR, FPR, FNR, FOR, FP cost, FN cost, and the overarching Total Cost.

\subsection{Data Preprocessing and Cleansing} \label{DataPreprocessingandCleansingSec}

\begin{figure}
    \centering
    \includegraphics{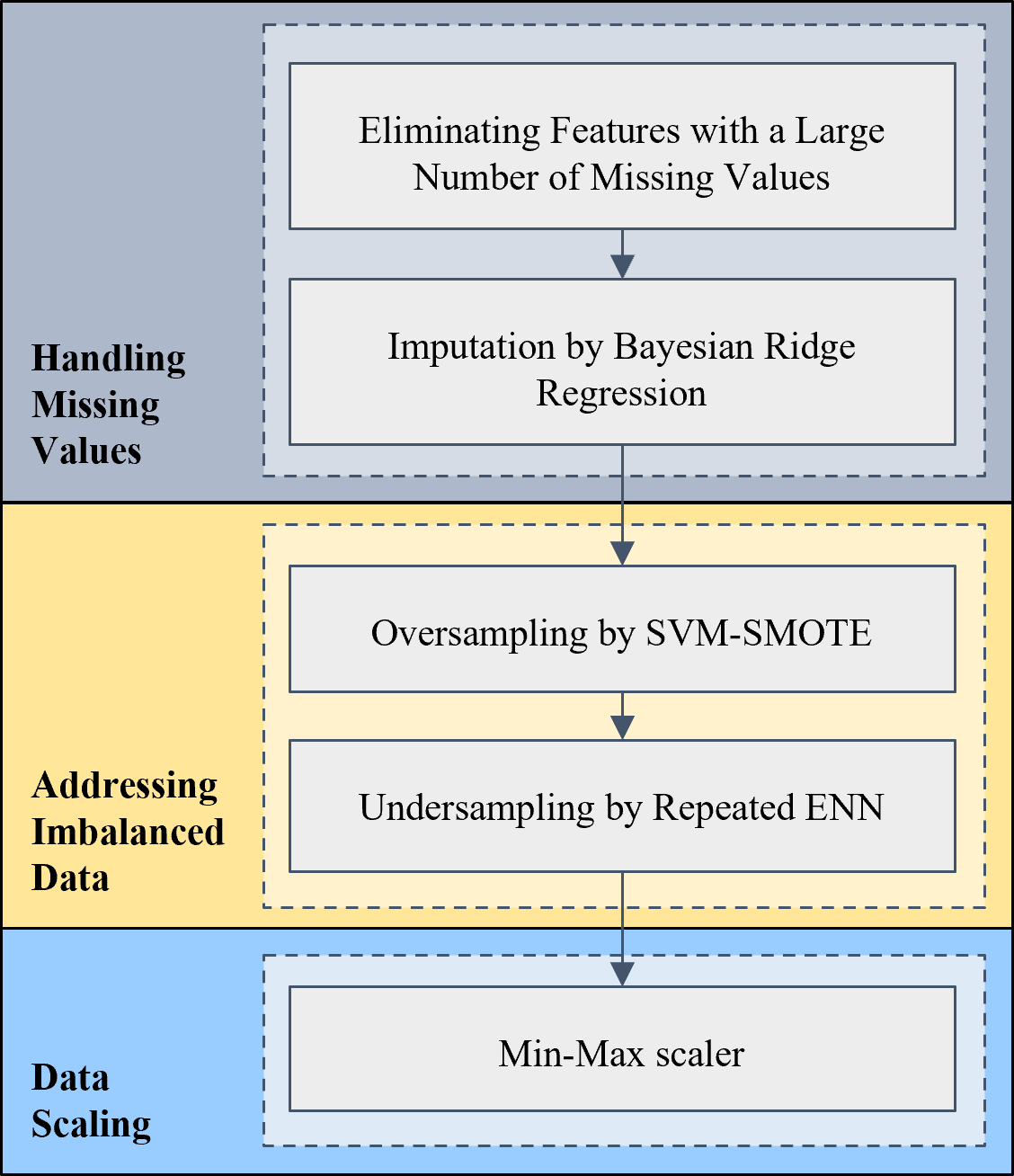}
    \caption{\rev{Our proposed data preprocessing workflow.}}
    \label{preprocessing}
\end{figure}

\begin{figure}
    \centering
    \includegraphics[width=\linewidth]{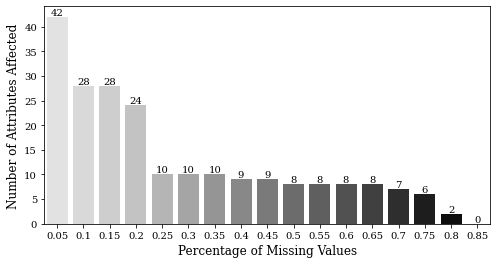}
    \caption{
    \rev{Exploring missing data patterns in the APS Failure at Scania Trucks Dataset during the data preparation process.}}
    \label{missing}
\end{figure}

\begin{figure}[ht] 
     \centering
     \subfloat[\rev{Original Training Set}]{
         \includegraphics[width=\linewidth]{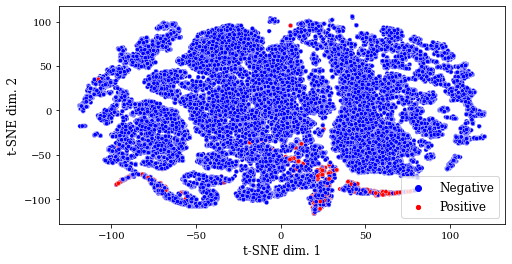}
         \label{tsnebefore}}
    \hfill
     \subfloat[\rev{Prepossessed Training Set}]{
         \includegraphics[width=\linewidth]{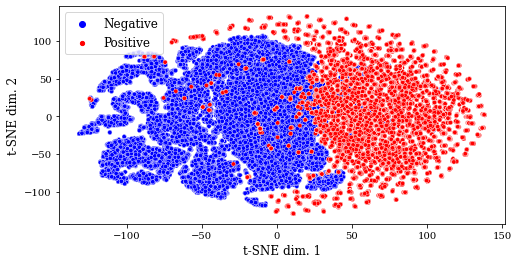}
         \label{tsneafter}}
        \caption{\rev{t-SNE projection of the APS Failure at Scania Trucks dataset.}
        }
        \label{tsne}
\end{figure}

\noindent In our data preparation process, illustrated in Fig. \ref{preprocessing}, one crucial step is the identification and handling of missing values.  These missing values arise from diverse sources, such as human errors, system glitches, or incomplete data collection. 
Addressing these gaps is essential to uphold the reliability and validity of statistical findings. 
We begin our data preparation process by assessing the extent to which various attributes were affected by missing data, as depicted in Fig. \ref{missing}. This analysis revealed that two features contained more than 80\% missing values, while twenty-eight attributes exhibited more than 10\% missing values. To streamline the dataset and reduce its dimensionality while preserving data integrity, we opted to eliminate attributes with more than 10\% missing values, resulting in the removal of a total of 28 attributes. This proactive approach aimed to mitigate noise and complexity, preventing biases or errors in our analysis.

In the second phase of data processing, we address the remaining incomplete data. Our approach involves utilizing Bayesian Ridge Regression for imputation, as detailed in Section \ref{metMisBay}. Initially, we make use of the mean as the starting point for imputing missing values. Then, we iteratively refine these estimates by integrating this regression model that leverages the available values of other features. This process is repeated for 100 iterations, ensuring a thorough and precise imputation of missing data. By adopting this meticulous approach, we enhance the dataset's completeness and reliability, setting the stage for subsequent analyses.

To tackle the challenge of class imbalance, the third and fourth phases of our data preprocessing workflow involve resampling techniques. As elucidated in Section \ref{metIm}, we employ a combination of oversampling and undersampling methods to effectively address the issue of imbalanced data. Specifically, our oversampling method of choice is SVM-SMOTE, which involves increasing the representation of the minority class until it reaches a balanced 50\% proportion compared to the majority class. Subsequently, we apply Repeated ENN as the undersampling technique to eliminate noisy samples from the majority class. As a result of this comprehensive class imbalance adjustment pipeline, our dataset consists of 57,136 negative samples and 29,500 positive samples. This remarkable improvement in the class imbalance ratio, from 59:1 to 1.94:1, contributes to a more equitable and robust dataset for the next analyses.

The final phase of our data processing pipeline involves scaling. A commonly employed normalization technique is known as Min-Max scaling. This method is characterized by the following equation:
\begin{equation}
    m = \frac{x-x_{min}}{x_{max}-x_{min}},
\end{equation}
Here, $m$ represents the scaled value, $x$ denotes the original value, and $x_{min}$ and $x_{max}$ stand for the minimum and maximum values of the respective feature. The Min-Max scaling approach serves to transform each feature within the dataset into a standardized range spanning from 0 to 1. In our research, we apply Min-Max scaling to ensure uniform scaling across all features. This not only enhances the performance and stability of algorithms sensitive to data scaling but also diminishes the impact of outliers and large values, contributing to more robust and consistent results.

To illustrate the effectiveness of our data preprocessing, we present t-SNE projections of the APS Failure at Scania Trucks dataset, comparing the original training set with the preprocessed one. t-SNE is a dimensionality reduction technique that enables us to visualize the distribution of data points in a lower-dimensional space while preserving their inherent relationships \cite{van2008visualizing}. As shown in Fig. \ref{tsne}, these visualizations showcase the transformative impact of our preprocessing steps, revealing a more favorable data distribution and improved class separability, highlighting the value of our approach in enhancing data quality and facilitating more robust analysis.

\subsection{Model Parameters and Configuration } \label{ModelParamSec}
\noindent In our endeavor to harness historical data for APS Failure detection, we implemented the Transformer model outlined in Section \ref{metModel} using Python 3.8.6. This computational task is executed on a computing server equipped with an NVIDIA Tesla A40 GPU boasting 48GB of RAM. 
We meticulously fine-tuned the hyperparameters, as detailed in Table \ref{hyperparameters}. The neural network weights were initialized using the default random initialization routines provided by the TensorFlow framework.

Given the stochastic nature inherent in neural networks, we conducted multiple experiments with five distinct random seeds for network initialization, thereby ensuring a fair and comprehensive comparison of results. During the model training process, we strategically allocated 10\% of the training set as a validation set, employing a stratified approach to maintain a balanced representation from both classes. 

Furthermore, we employed the Adam optimizer with a specific learning rate set at 0.0005. The batch size was configured to be 72, and the model training process ran for a maximum of 8000 epochs. 
However, during the training process, we diligently monitored model performance by implementing a checkpoint mechanism. This procedure involved saving the model weights that minimized the cost function, as defined in Equation \eqref{cost}, on the validation set. This meticulous approach allowed us to capture and retain the model's optimal state, ensuring that it continued to perform at its best on the test set.

\begin{table}[ht]
\centering
\caption{\rev{Our proposed model hyperparameters on the APS Failure at Scania Trucks dataset.}}
\label{hyperparameters}
\rev{\begin{tabular}{|c|c|c|c|}
\hline
\multicolumn{4}{|c|}{\textbf{Transformer input \& output sizes}} \\
\hline
Inputs & 142 & Outputs & 1 \\
\hline
\multicolumn{4}{|c|}{\textbf{Transformer encoder parameters}} \\
\hline
Number of transformer blocks & 4 & Dropout probability & 0.25 \\
\hline
Number of attention heads & 4 & Size of each attention  & 256\\
&& head for query and key & \\
\hline
\multicolumn{4}{|c|}{\textbf{Feed forward network parameters}} \\
\hline
Number of output filters & 4 & Dropout probability & 0.25 \\
\hline
\multicolumn{4}{|c|}{\textbf{MLP parameters}} \\
\hline
Size of dense layer 1&  128 & Dropout probability & 0.4 \\
and dense layer 2 & 64 &  &  \\
\hline
\multicolumn{4}{|c|}{\textbf{Focal Loss parameters}} \\
\hline
$\gamma$ & 1.5 & $\alpha$ & 0.95 \\
\hline
\end{tabular}}
\end{table}

\subsection{Results and Discussion} \label{ResultsandDiscussionSec}

\begin{figure}
    \centering
    \includegraphics[width=\linewidth]{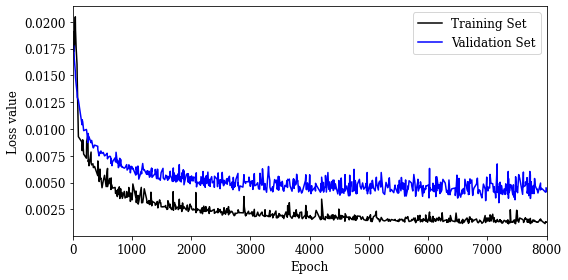}
    \caption{\rev{Learning curve of our Transformer based proposed model on the APS Failure at Scania Trucks dataset.}}
    \label{curve}
\end{figure}

\begin{figure}[ht] 
     \centering
     \subfloat[Train set]{
         \includegraphics[width=0.472\linewidth]{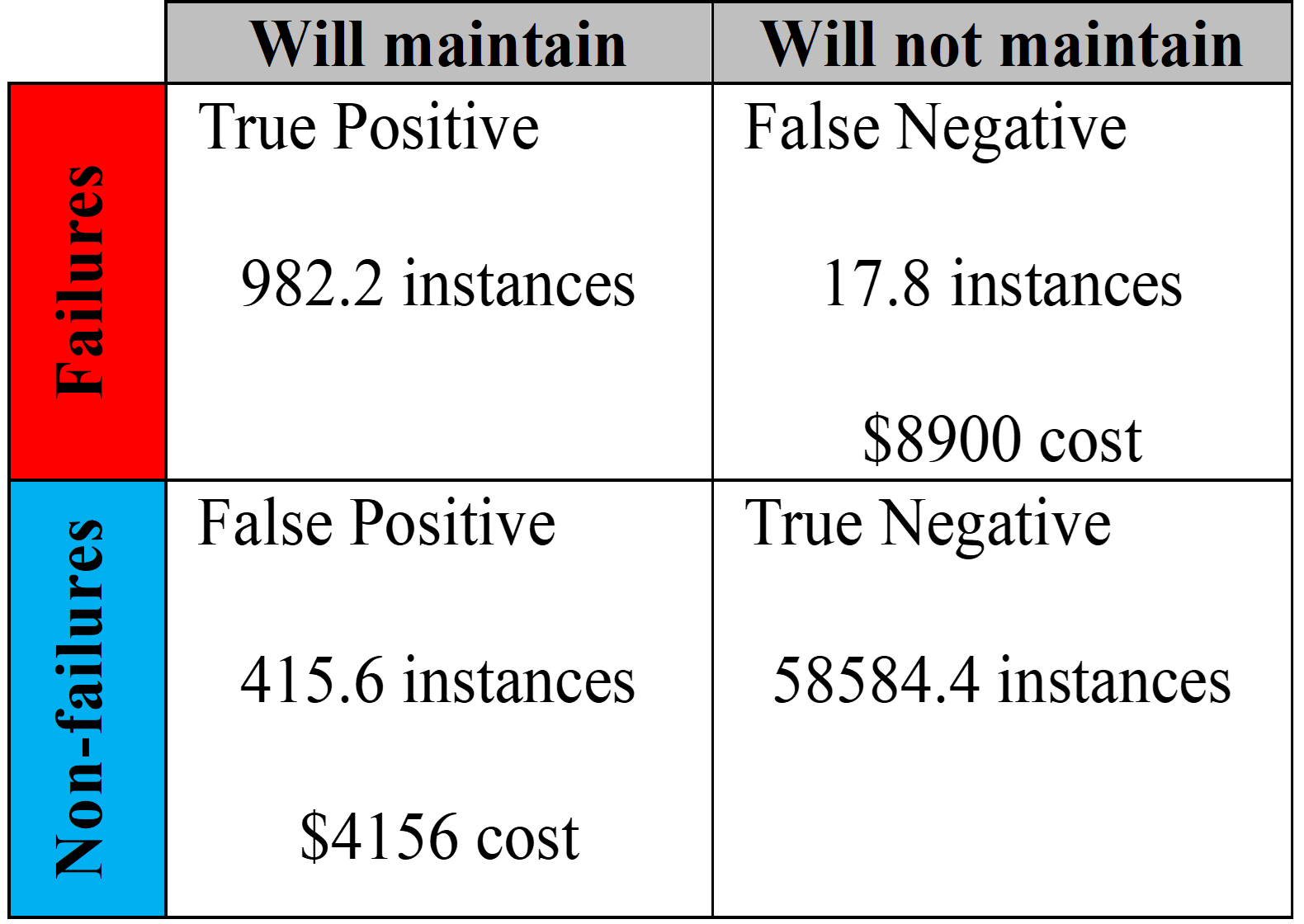}
         \label{cmtrain}}
     \subfloat[Test set]{
         \includegraphics[width=0.472\linewidth]{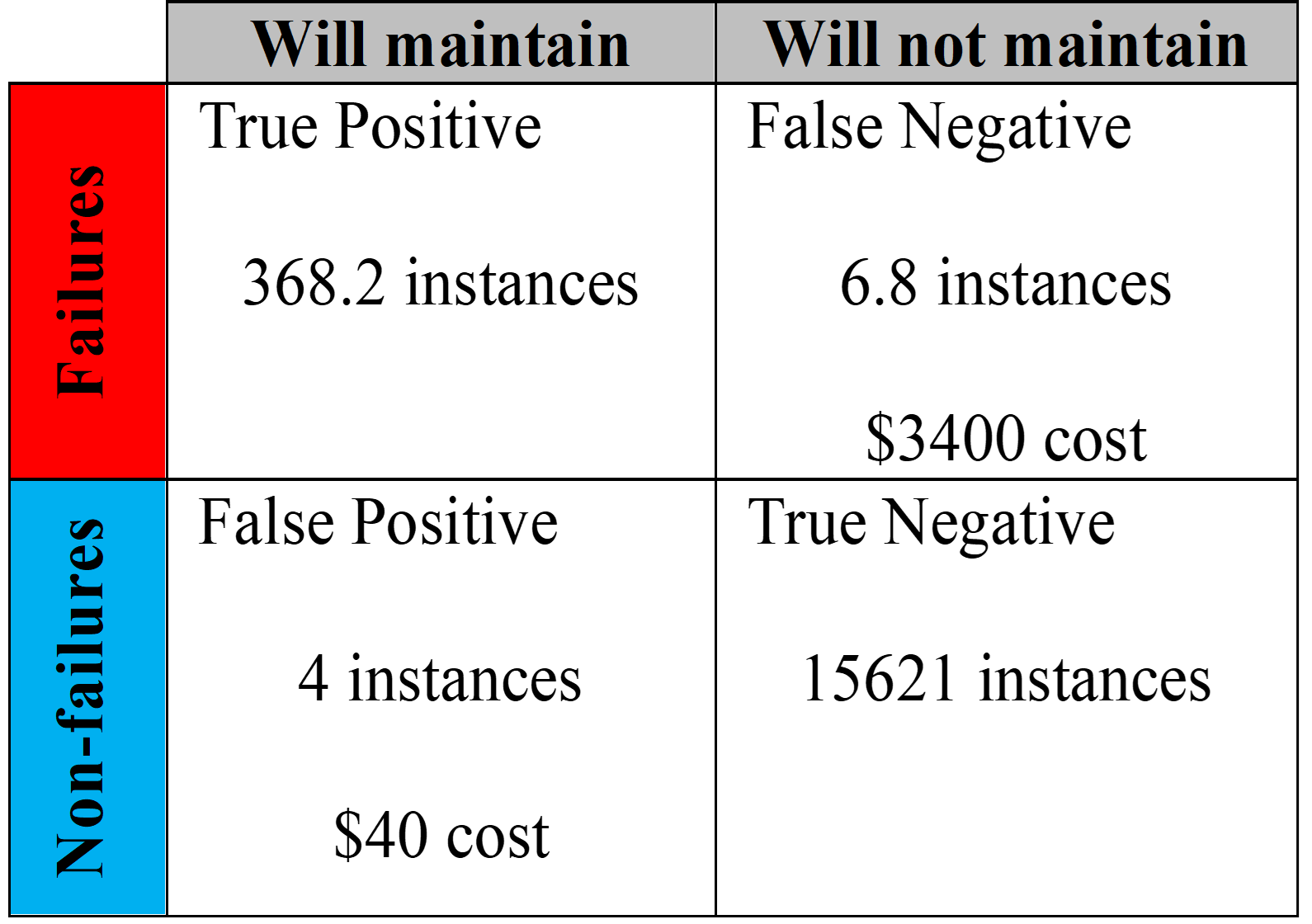}
         \label{cmtest}}
        \caption{Reliability-based confusion matrices providing a comprehensive overview of the performance of our Transformer model, tailored for cost-sensitive failure detection, on the APS Failure at Scania Trucks dataset.}
        \label{cm}
\end{figure}

\subsubsection{Learning Curve Analysis} Our model's learning curve is shown in Fig. \ref{curve}. This plot provides valuable insights into the training process, demonstrating that our model undergoes successful training without exhibiting signs of overfitting or underfitting, ensuring robust performance.

\subsubsection{Reliability-based Confusion Matrices} To comprehensively evaluate our Transformer model's performance, we employ confusion matrices, as illustrated in Fig. \ref{cm}, to have an aggregated perspective on our model's efficacy, customized for cost-sensitive failure detection. Importantly, we average these results over five different random seeds to demonstrate the stability of our approach.
On average, our model exhibits 17.8 FN instances, resulting in a cost of \$8,900 on the training set and 6.8 FN instances, amounting to a cost of \$3,400 on the test set. Additionally, we encounter a mere 4 FP instances on average in the test set, incurring a cost of \$40, totaling \$3,440.

\rev{
\subsubsection{Comparison with Prior Research} \label{ComparisonwithPriorResearchSec}
In Table \ref{tableCom}, we present an in-depth examination of models and strategies employed in recent research papers on the APS Failure at Scania Trucks dataset. Notably, our proposed method stands out as the only work that introduces a Transformer-based model to the APS failure prediction landscape. In addressing cost sensitivity, our innovation extends further, as we are the sole contributors to the design of a novel loss function, Focal Loss, showcasing our dedication to optimizing the cost.
In the realm of imbalanced data, our approach distinguishes itself by incorporating both oversampling (SVM-SMOTE) and undersampling (Repeated ENN), a unique combination that sets our work apart from others. Additionally, we address the challenge of missing values by using regression-based imputation.}

\rev{
We also provide a comparative analysis of our model's performance with that of other relevant research papers on the APS Failure at Scania Trucks dataset in Table \ref{table1} and Table \ref{table2}. We have reported our results on both the training and test sets while noting that most of the other papers do not report their results on the training set. 
In terms of training set performance, our model outperforms all others across all metrics. 
On the test set, we excel in terms of total cost, highlighting the successful design and superior generalization capabilities of our cost-sensitive Transformer-based model. 
We also rank among the top two performers in most other metrics. Notably, some previous works focus exclusively on minimizing FN \cite{38ranasinghe2019methodology, 41sun2023robust} or FP \cite{31oh2023quantum}.
}

\begin{table*}[t]
\caption{\rev{Comparison of models and strategies for handling missing values, imbalanced data, and cost sensitivity in recent studies on the APS failure dataset.}}
\label{tableCom}
\rev{
\centering
\begin{tabular}{|P{3.1cm} |P{4.1cm} |P{3.7cm} |P{1.8cm} |P{3.2cm}|}
\hline
Study Reference &	Handling Missing Values	&Addressing Imbalanced Data&	Addressing Cost Sensitivity&	Model \\
\hline
K. Taghandiki et al. \cite{36taghandiki2023minimizing}&	Mean imputation&	-&	-	&DNN \\
\hline
E. Oh et al. \cite{37oh2020imbalanced}& 	GPR-based estimation	&GAN-based oversampling	&-	&DNN \\
\hline
M. N. Syed et al. \cite{33syed2020novel}&	KNN-based imputation	&AUC maximization	&-	&Custom Linear Classifier \\
\hline
Q. Ke et al. \cite{27ke2022deep}&	-	&-	&-	&LSTM \\
\hline
S. Rafsunjani et al. \cite{30rafsunjani2019empirical} (The Best Result)&	Eliminating features with high missing values \& mean imputation&	Random undersampling	&-	&Random Forest \\
\hline
M. M. Akarte et al. \cite{40akarte2018predictive}&	Eliminating features with  high missing values \& median imputation&More class weight to the minority class &-	&XGBoost \\
\hline
K. Sun et al. \cite{41sun2023robust}&	Zero imputation &	Random undersampling	&-	&Siamese Neural Network \\
\hline
G. D. Ranasingh et al. \cite{38ranasinghe2019methodology} &	-	&CGAN-based oversampling	&-	&Random Forest \\
\hline
G. D. Ranasingh et al. \cite{38ranasinghe2019methodology} &	-	&CGAN-based oversampling	&-	&Gradient Boosting Machine \\
\hline
E. Oh et al. \cite{31oh2023quantum}	&GPR-based estimation&	Quantum mechanics-based oversampling&	-	&LSTM \\
\hline
\textbf{Proposed method}	& \textbf{Eliminating features with high missing values \& Bayesian Ridge Regression-based imputation} &	\textbf{SVM-SMOTE \& Repeated ENN}	 &\textbf{Focal Loss} &	\textbf{Transformer} \\
\hline
\end{tabular}}
\end{table*}

\rev{
To be more specific, our competitor, \cite{31oh2023quantum}, exhibits certain limitations that warrant careful consideration. 
First, as mentioned, a notable divergence lies in the primary focus of \cite{31oh2023quantum}, which centers on minimizing FP. While this approach can yield favorable results in metrics dependent solely on FP, it introduces challenges in scenarios where the cost associated with FN holds greater significance.
In contrast, our model distinguishes itself by incorporating Focal Loss, a strategic choice that allows us to adapt to varying cost structures. Unlike \cite{31oh2023quantum}, our approach does not solely concentrate on minimizing FP but rather maintains a balance between FN and FP. This adaptability proves particularly advantageous in scenarios with dynamically changing costs, ensuring that our model remains robust and cost-effective across a spectrum of real-world applications.
Furthermore, their approach is encumbered by a high computational cost, potentially hindering its practicality in real-world applications. 
Moreover, the lack of available code raises concerns about the replicability and credibility of their findings, a critical aspect for fostering transparency and further research (see more details in Appendix).
}

\rev{ 
\subsubsection{Ablation Study} \label{AblationStudySec}
In the conducted ablation study, as shown in Table \ref{tableAb}, we systematically evaluated the impact of different components in our proposed method for APS failure prediction. 
In the case of \textit{No Eliminating Features}, the minimal difference observed suggests that retaining all features does not significantly contribute to performance in this context. However, this outcome may be attributed to the heightened challenge of finding a robust imputer when a substantial portion of the features has missing values. By opting to eliminate features with high missing values, not only do we conserve computational resources, but we also mitigate noise in the dataset for the imputer. This strategic removal enhances efficiency by providing a less noisy and more manageable dataset for the imputation process.}

\rev{
On the flip side, the absence of oversampling, as indicated by \textit{No Oversampling}, leads to a decline in performance. This underscores the critical role of addressing the class imbalance, as the deterioration observed highlights the ongoing challenge posed by imbalanced data when SVM-SMOTE is omitted.
The scenario of \textit{No Undersampling} yields undesirable outcomes attributed to the sensitivity of Focal Loss to outliers. This sensitivity causes Focal Loss to prioritize outliers as hard examples, diverting attention away from establishing a robust decision boundary for the majority class. Consequently, this shift in focus contributes to an elevated number of FP.}

\rev{
Finally, our experiments reveal that Focal Loss emerges as the most crucial component, as evidenced by the worst results obtained when employing a cross entropy loss instead. Without Focal Loss, the model loses control over the cost-sensitive aspects, leading to increased FN and, consequently, higher costs. This underscores the pivotal nature of Focal Loss in optimizing the cost rather than 
FP and FN.
Moreover, it is important to acknowledge the substantial role played by the Transformer model, contributing to the overall performance.
}


\rev{
A further instance showcasing the applicability of our approach is presented in Appendix, where we delve into a case study focused on identifying defective-manufactured semiconductors.
}

\begin{table*}[ht]
\centering
\caption{Comparison of APS failure prediction performances reported in recent studies on the APS failure dataset.}
\label{table1}
\begin{tabular}{|c|c|c|c|c|c|c|c|c|c|c|}
\hline
\multicolumn{11}{|c|}{\textbf{Train set}} \\
\hline
Study Reference & TP & FP & FN & TN & Accuracy & Precision & Sensitivity (Recall) & Specificity & F1 Score & NPV \\
\hline
M. N. Syed et al. \cite{33syed2020novel} & 967 & 1568 & 33 & 57432 & 0.9733 & 0.3815 & 0.967 & 0.9734 & 0.5471 & 0.9994 \\
\hline
\textbf{Proposed Method} & \textbf{982.2} & \textbf{415.6} & \textbf{17.8} & \textbf{58584.4} & \textbf{0.9928} & \textbf{0.7027} & \textbf{0.9822} & \textbf{0.9930} & \textbf{0.8183} & \textbf{0.9997} \\
\hline
\hline
\multicolumn{11}{|c|}{\textbf{Test set}} \\
\hline
K. Taghandiki et al. \cite{36taghandiki2023minimizing} & 238 & 77 & 137 & 15548 & 0.9866 & 0.7556 & 0.6347 & 0.9951 & 0.6899 & 0.9913 \\
E. Oh et al. \cite{37oh2020imbalanced} & 316 & 207 & 59 & 15418 & 0.983 & 0.6042 & 0.8427 & 0.9868 & 0.7038 & 0.9962 \\
M. N. Syed et al. \cite{33syed2020novel} & 345 & 519 & 30 & 15106 & 0.96569 & 0.3993 & 0.92 & 0.9668 & 0.5569 & 0.998 \\
Q. Ke et al. \cite{27ke2022deep} & 347 & 229 & 28 & 15396 & 0.9839 & 0.6024 & 0.9253 & 0.9853 & 0.7298 & 0.9982 \\
S. Rafsunjani et al. \cite{30rafsunjani2019empirical} & & & & & & & & & & \\ 
(The Best Result) & 366 & 771 & 9 & 14854 & 0.9513 & 0.3219 & 0.976 & 0.9507 & 0.4841 & 0.9994 \\
M. M. Akarte et al. \cite{40akarte2018predictive} & 363 & 414 & 12 & 15211 & 0.9734 & 0.4672 & 0.968 & 0.9735 & 0.6302 & 0.9992 \\
K. Sun et al. \cite{41sun2023robust} & \textbf{369} & 608 & \textbf{6} & 15017 & 0.9616 & 0.3777 & 0.984 & 0.9611 & 0.5459 & \textbf{0.9996} \\
G. D. Ranasingh et al. \cite{38ranasinghe2019methodology} & & & & & & & & & & \\ 
(Random Forest) & \textbf{371} & 405 & \textbf{4} & 15220 & 0.9744 & 0.4781 & \textbf{0.9893} & 0.9741 & 0.6447 & \textbf{0.9997} \\
G. D. Ranasingh et al. \cite{38ranasinghe2019methodology} & & & & & & & & & & \\ 
(Gradient Boosting Machine) & \textbf{369} & 255 & \textbf{6} & 15370 & 0.9837 & 0.5913 & \textbf{0.984} & 0.9837 & 0.7387 & \textbf{0.9996} \\
E. Oh et al. \cite{31oh2023quantum} & 367 & \textbf{0} & 8 & \textbf{15625} & \textbf{0.9995} & \textbf{1} & 0.9787 & \textbf{1} & \textbf{0.9892} & 0.9995 \\
\hline
\textbf{Proposed Method} & 368.2 & \textbf{4} & 6.8 & \textbf{15621} & \textbf{0.9993} & \textbf{0.9892} & 0.9818 & \textbf{0.9997} & \textbf{0.9856} & \textbf{0.9996} \\
\hline
\end{tabular}
\end{table*}

\begin{table*}[ht]
\centering
\caption{Comparison of APS failure prediction performances reported in recent studies on the APS failure dataset.}
\label{table2}
\begin{tabular}{|c|c|c|c|c|c|c|c|}
\hline
\multicolumn{8}{|c|}{\textbf{Train set}} \\
\hline
Study Reference & FDR &FPR (Fall Out) & FNR & FOR & FP Cost in \$ & FN Cost in \$  & Total Cost in \$  \\
\hline
M. N. Syed et al. \cite{33syed2020novel} & 0.6185 & 0.0266 & 0.033 & 0.0006 & 15680 & 16500 & 32180 \\
\hline
\textbf{Proposed Method}  & \textbf{0.2973} & \textbf{0.0070} & \textbf{0.0178} & \textbf{0.0003} & \textbf{4156} & \textbf{8900} & \textbf{13056} \\
\hline
\hline
\multicolumn{8}{|c|}{\textbf{Test set}} \\
\hline
K. Taghandiki et al. \cite{36taghandiki2023minimizing} & 0.2444 & 0.0049 & 0.3653 & 0.0087 & 770 & 68500 & 69270 \\
E. Oh et al. \cite{37oh2020imbalanced} & 0.3958 & 0.0132 & 0.1573 & 0.0038 & 2070 & 29500 & 31570 \\
M. N. Syed et al. \cite{33syed2020novel}  & 0.6007 & 0.0332 & 0.08 & 0.0020 & 5190 & 15000 & 20190 \\
Q. Ke et al. \cite{27ke2022deep} & 0.3976 & 0.0147 & 0.0747 & 0.0018 & 2290 & 14000 & 16290 \\
S. Rafsunjani et al. \cite{30rafsunjani2019empirical} & & & & & & & \\
(The Best Result) & 0.6781 & 0.0493 & 0.024 & 0.0006 & 7710 & 4500 & 12210 \\
M. M. Akarte et al. \cite{40akarte2018predictive} & 0.5328 & 0.0265 & 0.032 & 0.0008 & 4140 & 6000 & 10140 \\
K. Sun et al. \cite{41sun2023robust} & 0.6223 & 0.0389 & 0.016 & \textbf{0.0004} & 6080 & \textbf{3000} & 9080 \\
G. D. Ranasingh et al. \cite{38ranasinghe2019methodology} & & & & & & & \\
(Random Forest) & 0.5219 & 0.0259 & \textbf{0.0107} & \textbf{0.0003} & 4050 & \textbf{2000} & 6050 \\
G. D. Ranasingh et al. \cite{38ranasinghe2019methodology} & & & & & & & \\
(Gradient Boosting Machine) & 0.4087 & 0.0163 & 0.016 & \textbf{0.0004} & 2550 & \textbf{3000} & 5550 \\
E. Oh et al. \cite{31oh2023quantum} & \textbf{0} & \textbf{0} & 0.0213 & 0.0005 & \textbf{0} & 4000 & 4000 \\
\hline
\textbf{Proposed Method} & \textbf{0.0108} & \textbf{0.0003} & \textbf{0.0187} & \textbf{0.0004}  & \textbf{40} & 3400 & \textbf{3440} \\
\hline
\end{tabular}
\end{table*}

\begin{table}[ht]
\centering
\caption{\rev{Ablation Study: Evaluating APS failure prediction performances with variations in our proposed method on the APS failure test dataset.}}
\label{tableAb}
\rev{\begin{tabular}{|c|c|c|c|c|c|}
\hline
Ablation Study & TP &FP & FN& TN & Cost in \$  \\
\hline
\textbf{Full Method} & \textbf{368.2} & \textbf{4} & \textbf{6.8} & \textbf{15621}  & \textbf{3440} \\
\hline
No Eliminating Features &	368 &	8	&7	&15617	&3580\\
No Oversampling &	364.6	&6	&10.4	&15619&	5260\\
No Undersampling&	367	&101	&8	&15524	&5010\\
No Focal Loss&	339&	82	&36	&15543	&18820\\
\hline
\end{tabular}}
\end{table}

\section{Conclusion} \label{conc}
\noindent \rev{In this study, we have harnessed the transformative capabilities of the Transformer architecture along with a cost-capturing loss function, introducing its application in the realm of predictive maintenance, with a specific emphasis on prognostics and failure detection. By leveraging the intrinsic power of Transformers, we have reshaped the landscape of protective maintenance, marking a substantial departure from traditional ML approaches in this domain. 
Our approach capitalizes on the self-attention mechanism and contextual learning abilities inherent in Transformers, resulting in improved predictive accuracy, early detection of potential failures, and, ultimately, the optimization of equipment reliability and operational efficiency.
Additionally, for the first time,  we introduced a hybrid resampling approach, further enhancing our method's effectiveness.
Our findings underscore the profound impact of our cost-sensitive Transformer-based model, illustrating its potential for real-world applications in APS failure detection scenarios, and detecting defective-manufactured semiconductors. Through rigorous testing on both APS and SECOM datasets, we have demonstrated a substantial enhancement in performance compared to state-of-the-art methods. Also, we conducted an ablation study, dissecting the contributions of different components in our proposed method.
As we look ahead, future work can explore the extension of our effective and data-driven approach to protective maintenance across various industrial sectors. By doing so, we can further advance the reliability and efficiency of industrial operations, ushering in a new era of predictive maintenance.}

\rev{
\appendix[Case Study \MakeUppercase{\romannumeral 2}: Detecting Defective-Manufactured Semiconductors]} \label{SECOMSec}

\section*{\rev{Problem Description \& SECOM Dataset}} \label{ProblemDescriptionSECOMSec}
\noindent \rev{The difficulties associated with imbalanced data, missing values, and cost sensitivity are not confined solely to the heavy-duty vehicle domain; they also manifest in the semiconductor manufacturing process. 
Manufacturing data is collected from various semiconductor manufacturing systems. Despite precise control of each device, small oscillating movements are generated during operations. For instance, a photolithography machine, responsible for inputting a circuit pattern, experiences fluctuations due to internal motor movements and mechanisms. Similar oscillations occur in subsequent processes, such as the metal lining machine. These vibrations contribute to the introduction of erroneous data through the sensing mechanisms attached to process equipment. 
Furthermore, the presence of missing values can be attributed to various factors, such as failures in routine maintenance activities or sudden malfunctions of sensing systems, leading to the generation of missing values. 
Detecting defective semiconductors involves handling a significantly smaller number of flawed chips compared to their defect-free counterparts, thereby posing a challenge related to class imbalance. 
Lastly, there is the additional concern of the asymmetric cost associated with failing to detect flawed chips.
In this context, the precise prediction of failures during the manufacturing process is becoming progressively indispensable.}

\rev{In this section, we employ our proposed method to predict equipment faults during the wafer fabrication process in semiconductor industries. We utilize the SECOM (Semiconductor Manufacturing) dataset as our source of data \cite{secom}.
This dataset consists of manufacturing operation data and semiconductor quality data, with 1567 observations taken from a wafer fabrication production line. Each observation is a vector of 590 sensor measurements, along with a label of pass/fail test.
There are only 104 fail cases labeled as positive, while a much larger number of examples pass the test and are labeled as negative. This results in a 14:1 proportion, posing a significant class imbalance challenge. 
}
\section*{\rev{Results on SECOM dataset}} \label{ResultsSECOMSec}
\noindent \rev{Firstly, we implement a stratified $K$-fold cross-validation with $K=8$ to segregate and generate training and test sets. For each fold, we ensure 13 positive samples in the test set and 91 in the training set. Subsequently, we apply the same data preprocessing workflow as illustrated in Fig. \ref{preprocessing}, akin to the procedures applied to the APS dataset in Section \ref{DataPreprocessingandCleansingSec}. Through this process, the number of attributes is reduced to 538 by eliminating 52 features out of the total 590. Following this, our resampling technique is applied, resulting in an improvement of the class imbalance to 1.37:1. Finally, the scaling is applied as the concluding stage of the preprocessing.}

\rev{
We utilize identical model parameters and configurations as detailed in Section \ref{ModelParamSec}, with the exception of the following adjustments: Input Sizes set to 538, the size of attention head for query and key set to 64, the number of transformer blocks reduced to 1, the number of attention heads set to 1, and the size of the dense layer set to 2.
}

\rev{
In Table \ref{tableSECOM}, we present a comparative analysis, evaluating the performance of our model in relation to other relevant papers on the SECOM dataset. The assessment focuses on crucial performance metrics outlined in Section \ref{PerformanceMetricsSec}, and the cost is computed using Equation \eqref{cost}. To ensure the robustness of our findings, we average the results over five different random seeds.
On average, our model demonstrates 1.4 instances of FN, incurring a cost of \$700, and 25 instances of FP, resulting in a cost of \$250. The total cost amounts to \$950. Notably, our model outperforms all others across a majority of metrics, showcasing its effective design and superior generalization capabilities.
On the flip side, it is pertinent to add to our discussion about the study \cite{31oh2023quantum} in Section \ref{ComparisonwithPriorResearchSec}, that there is a discrepancy in the reported confusion matrix. Specifically, the indicated data size is 1671, whereas it should be 1567. This inconsistency raises concerns about the accuracy of their reported results and warrants further examination. 
}

\begin{table}[ht]
\centering
\caption{\rev{Comparison of failure prediction performances reported in recent studies on the SECOM dataset.}}
\label{tableSECOM}
\rev{\begin{tabular}{|c|c|c|c|c|c|c|}
\hline
Study Reference & Data & TP &FP & FN& TN & Total   \\
 & Size &  & & &  &  Cost in \$  \\
\hline
E. Oh et al. \cite{37oh2020imbalanced} &1671&	96&	110	&8	&1457	&5100 \\ 
E. Oh et al. \cite{31oh2023quantum}& 1671&	102&	0	&2	&1567&	1000 \\
\hline
\textbf{Proposed Method}	& 1567 &\textbf{102.6}	& 25 &	\textbf{1.4}&	\textbf{1438}	&\textbf{950} \\
\hline
\end{tabular}}
\end{table}

\balance
\bibliographystyle{IEEEtran}
\bibliography{ref.bib}
\end{document}